\documentclass{egpubl}
\usepackage{eurovis2023}

\SpecialIssuePaper
\CGFccby

\usepackage[T1]{fontenc}
\usepackage{dfadobe}  

\biberVersion
\BibtexOrBiblatex

\electronicVersion
\PrintedOrElectronic

\usepackage[pdftex]{graphicx}
\usepackage{egweblnk}

\usepackage[
    backend=biber,
    bibstyle=EG,
    citestyle=alphabetic,
    backref=true,
    giveninits=true,
    ]
    {biblatex}

\usepackage[scale=0.92]{inconsolata}
\usepackage{multicol}
\usepackage{listings}
\usepackage[most]{tcolorbox}
\usepackage{xcolor}
\usepackage{csquotes}
\usepackage{siunitx}
\usepackage{booktabs}

\DeclareMathOperator{\cov}{cov}

\definecolor{pd-petrol}{HTML}{1c5c60}
\definecolor{pd-aqua}{HTML}{23c7a2}
\definecolor{pd-yellow}{HTML}{ffcc00}
\definecolor{pd-red}{HTML}{ed3d3d}
\definecolor{pd-pink}{HTML}{f7a3bf}
\definecolor{pd-brown}{HTML}{9d6700}
\definecolor{pd-grey}{HTML}{91907e}
\definecolor{pd-green}{HTML}{579d04}
\definecolor{pd-purple}{HTML}{a35c80}
\definecolor{pd-orange}{HTML}{fd771e}

\colorlet{digit}{pd-green}
\colorlet{key}{pd-petrol}
\colorlet{value}{pd-grey!20!black}

\lstdefinelanguage{yaml}{
  aboveskip=-2.5ex,
  belowskip=-.25ex,
  xleftmargin=0pt,
  xrightmargin=01pt,
  framexleftmargin=5pt,
  framexrightmargin=5pt,
  keywords={},
  keywordstyle=\bfseries\color{key},
  basicstyle=\footnotesize\ttfamily\color{value},
  sensitive=false,
  comment=[l]{\#},
  morecomment=[s]{/*}{*/},
  numberstyle=\color{pd-red},
  commentstyle=\color{pd-brown},
  moredelim=[is][\bfseries\color{key}]{~}{~},
  moredelim=[s][\color{pd-grey}]{<}{>},
  literate=%
    *{0}{{{\color{digit}0}}}1
     {1}{{{\color{digit}1}}}1
     {2}{{{\color{digit}2}}}1
     {3}{{{\color{digit}3}}}1
     {4}{{{\color{digit}4}}}1
     {5}{{{\color{digit}5}}}1
     {6}{{{\color{digit}6}}}1
     {7}{{{\color{digit}7}}}1
     {8}{{{\color{digit}8}}}1
     {9}{{{\color{digit}9}}}1
     {...}{{{\color{pd-grey!70!white}...}}}3,
  classoffset=1,
  otherkeywords={-,:,\{,\},[,]},
  morekeywords={-,:,\{,\},[,]},
  keywordstyle=\bfseries\color{pd-aqua},
  classoffset=0,
}

\newtcblisting{spec}{
    boxsep=0pt,
    left=5pt,
    top=15pt,
    bottom=5pt,
    colback=pd-brown!5!white,
    boxrule=0pt,
    frame empty,
    arc=0pt,
    listing only,
    listing options={
        language=yaml,
    }
}

\newtcblisting{spectc}{
    enhanced,
    boxsep=0pt,
    left=5pt,
    top=3pt,
    bottom=5pt,
    colback=pd-brown!5!white,
    boxrule=0pt,
    frame empty,
    arc=0pt,
    listing only,
    listing options={
        language=yaml,
        multicols={2},
        escapeinside={`}{`},
    },
    overlay = {\draw[line width=1mm, white] (frame.north) -- (frame.south);}
}    

\newcommand{\gkey}[1]{\texttt{\bfseries\color{key}#1}}
\newcommand{\gval}[1]{\texttt{\color{value}#1}}

\title[ParaDime]{ParaDime: A~Framework for Parametric Dimensionality Reduction}

\author[A. Hinterreiter et al.]{%
    \parbox{\textwidth}{\centering%
        Andreas Hinterreiter%
        \textsuperscript{1}%
        \orcid{0000-0003-4101-5180},
        Christina Humer%
        \textsuperscript{1}%
        \orcid{0000-0002-0249-4062},
        Bernhard Kainz%
        \textsuperscript{2,3}%
        \orcid{0000-0002-7813-5023}, and
        Marc Streit%
        \textsuperscript{1}%
        \orcid{0000-0001-9186-2092}%
    }\\
    {\parbox{\textwidth}{\centering%
        \textsuperscript{1}%
        Johannes Kepler University Linz, Austria\\
        \textsuperscript{2}%
        Friedrich-Alexander-University Erlangen-Nuremberg, Germany\\
        \textsuperscript{3}%
        Imperial College London, UK%
    }}
}

\begin{document}

\teaser{%
    \vspace*{-0.5cm}
    \includegraphics[width=\linewidth]{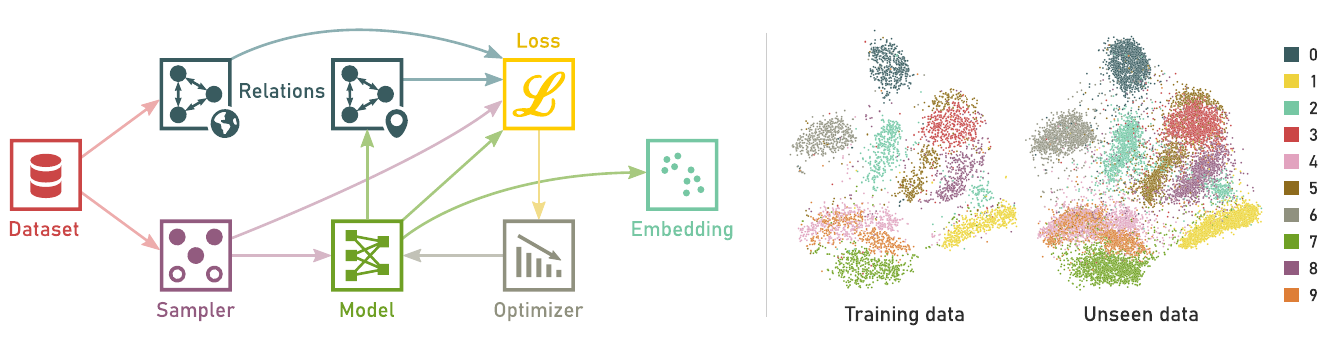}
    \centering
    \caption{ParaDime is a framework for parametric dimensionality reduction. Left:~Data flow in a single training phase of a ParaDime routine. Right:~Parametric \textit{t}-SNE trained on a subset of 5000 images from the MNIST dataset~\cite{lecun_mnist_2005} and applied to 15,000 unseen images.}
    \label{fig:teaser}
}

\maketitle

\begin{abstract}
ParaDime is a framework for parametric dimensionality reduction~(DR). In parametric DR, neural networks are trained to embed high-dimensional data items in a low-dimensional space while minimizing an objective function. ParaDime builds on the idea that the objective functions of several modern DR techniques result from transformed inter-item relationships. It provides a common interface for specifying these relations and transformations and for defining how they are used within the losses that govern the training process. Through this interface, ParaDime unifies parametric versions of DR techniques such as metric MDS, \textit{t}-SNE, and UMAP. It allows users to fully customize all aspects of the DR process. We show how this ease of customization makes ParaDime suitable for experimenting with interesting techniques such as hybrid classification\slash embedding models and supervised DR. This way, ParaDime opens up new possibilities for visualizing high-dimensional data.



\end{abstract}  


\section{Introduction}

Dimensionality reduction (DR) is one of the standard strategies for visualizing high-dimensional data.
The general concepts of DR have been known and applied for over a century~\cite{abdi_principal_2010} in the form of linear techniques such as principal component analysis~(PCA).
In recent decades, however, nonlinear DR techniques have gained popularity. The most prominent modern techniques are t-distributed stochastic neighbor embedding~(\textit{t}-SNE)~\cite{van_der_maaten_visualizing_2008} and uniform manifold approximation and projection~(UMAP)~\cite{mcinnes_umap:_2018}.
Both \textit{t}-SNE and UMAP rely on pairwise inter-item relationship information from  high-dimensional data to construct embeddings in a low-dimensional space, with the goal of preserving key \enquote{structures} of the original data.

One shortcoming of such relationship-based DR techniques is that new items cannot readily be added to existing embeddings without recomputing all pairwise relationships.
To address this shortcoming, researchers have developed \emph{parametric} DR techniques.
In parametric DR, embeddings are created by parameterized functions (e.g., neural networks) that are trained on high-dimensional data.
While several implementations of parametric DR are available, most of them are tailor-made variations of existing techniques, and they are often difficult to customize or extend.
This \enquote{scattered} nature of existing parametric DR techniques is surprising, considering the strong conceptual similarities between various nonlinear DR approaches~\cite{bohm_attraction_2022}.
In particular, the loss in many DR techniques is based on the comparison of (transformed) pairwise distances between data points.

We believe that the potential of parametric DR is underexplored, and that both the visualization and the machine learning communities could benefit from a framework that makes it easy to experiment with such techniques.
For this purpose, we introduce \emph{ParaDime}, a unifying framework for parametric DR.
Our contribution with ParaDime is threefold:
\begin{itemize}
    \item We introduce a generalizing grammar which formalizes all the steps and building blocks necessary to specify a parametric DR routine.
    \item We show how this grammar can be used not only to create parametric versions of existing DR techniques, but also to experiment with new ideas.
    \item We present an implementation of the grammar with a focus on usability and customization.
\end{itemize}

Our paper is structured as follows:
In Section~\ref{sec:related-work} we summarize the historical development of DR, focusing on parametric techniques; in Section~\ref{sec:grammar} we explain how similarities between these techniques give rise to a grammar of parametric DR, and how ParaDime implements this grammar;
we then show how ParaDime can be used to create parametric versions of existing DR techniques (Section~\ref{sec:existing-techniques}) and how it facilitates experimentation with new ideas (Section~\ref{sec:new-ideas});
in Section~\ref{sec:discussion} we discuss design choices, ease of use, limitations, and future work;
Section~\ref{sec:conclusion} concludes the paper.

\section{Related Work}
\label{sec:related-work}

DR can be categorized broadly into linear and nonlinear techniques.
The oldest linear technique, PCA, has been known for over a century~\cite{pearson_lines_1901,abdi_principal_2010}.
The survey by Cunningham and Ghahramani~\cite{cunningham_linear_2015} provides an excellent overview of the many linear techniques that have been developed since PCA was introduced.
Among these, multidimensional scaling~(MDS)~\cite{torgerson_multidimensional_1952} is most pertinent to our work.
Classic MDS is an eigenvalue problem with a close relationship to PCA~\cite{williams_connection_2002,chen_multidimensional_2008}.
In contrast, \emph{metric} MDS is a more general approach that aims to find a low-dimensional configuration of points whose pairwise distances best match those of the high-dimensional data.

Metric MDS, with its principle of comparing pairwise distances, is the intellectual  predecessor of many modern nonlinear techniques, such as Isomap~\cite{tenenbaum_global_2000}, SNE~\cite{hinton_stochastic_2002}, \textit{t}-SNE~\cite{van_der_maaten_visualizing_2008}, and UMAP~\cite{mcinnes_umap:_2018}.
Isomap tries to find a low-dimensional configuration based on geodesic (i.e., shortest-path) distances computed on a high-dimensional neighbor graph~\cite{tenenbaum_global_2000}.
In SNE, Gaussian kernels are used to transform pairwise distances into neighborhood probability distributions for  both the high- and low-dimensional data.
These probability distributions are then compared using the Kullback--Leibler~(KL) divergence~\cite{hinton_stochastic_2002}. 
To avoid the so-called crowding problem in the resultant embeddings, \textit{t}-SNE computes the probabilities in the low-dimensional space using the more fat-tailed Student's t-distribution~\cite{van_der_maaten_visualizing_2008}.
Finally, UMAP replaces the t-distribution with a modified Cauchy distribution and uses a cross entropy loss instead of the KL divergence~\cite{mcinnes_umap:_2018,sainburg_parametric_2021}.
The conceptual similarities of these (and several more) nonlinear DR techniques were highlighted in various contexts by Bengio et al.~\cite{bengio_learning_2004}, Böhm et al.~\cite{bohm_attraction_2022}, and Agrawal et al.~\cite{agrawal_minimum-distortion_2021}.
Recently, the relationship between \textit{t}-SNE and UMAP has been the subject of intense debate~\cite{becht_dimensionality_2019,kobak_initialization_2021,damrich_umaps_2021,damrich_contrastive_2022}.

Aside from their conceptual similarities, many nonlinear DR techniques share a practical limitation: they involve the calculation of pairwise distances.
Adding new points to existing embeddings---a problem known as out-of-sample extension---usually requires recomputing the whole embedding.
This drawback has been addressed by approximating embeddings with parametric functions, using (\textit{i})~kernel-based approaches~\cite{bengio_learning_2004,gisbrecht_linear_2012,gisbrecht_parametric_2015}, (\textit{ii})~mixture models and data imputation~\cite{de_bodt_nonlinear_2019}, or (\textit{iii})~neural networks~\cite{van_der_maaten_learning_2009,min_deep_2010,sainburg_parametric_2021,lai_parametric_2022}.
Parametric versions of \textit{t}-SNE and UMAP are examples of manifold learning~\cite{bengio_representation_2013}, a subfield of representation learning.
The general idea of using neural networks to reduce data dimensionality, in particular with autoencoders, predates these extensions~\cite{hinton_autoencoders_1993,hinton_reducing_2006}.
Additionally, parametric nonlinear DR techniques based on neighborhood information are related to metric learning~\cite{kulis_metric_2012}, where representations are determined by learning a distance function.

Minimum distortion embeddings~(MDEs)~\cite{agrawal_minimum-distortion_2021} and the matrix optimization framework by Cunningham and Ghahramani~\cite{cunningham_linear_2015} are closely related to our work
in that they aim to unify several existing techniques in a common framework.
Cunningham and Ghahramani view DR as a matrix optimization problem with varying objectives~\cite{cunningham_linear_2015}. By choosing the right objective and/or matrix constraints, a wide variety of techniques can be expressed in their framework---albeit only \emph{linear} ones. MDEs use formalized distortions and penalty functions to generalize non-linear embeddings~\cite{agrawal_minimum-distortion_2021}.
However, MDEs are non-parametric and support out-of-sample-extension only via a combination of anchoring constraints and solving a new MDE sub-problem~\cite{agrawal_minimum-distortion_2021}.
In addition,  MDEs are phrased in a way that makes it challenging to map them to existing techniques (see, e.g., the comparison of \textit{t}-SNE and UMAP by Sainburg et al.~\cite{sainburg_parametric_2021} vs.~how Agarwal et al.~relate penalties to UMAP~\cite{agrawal_minimum-distortion_2021}).
ParaDime focuses instead on (potentially transformed) pairwise relations between data items, which allows several existing techniques to be directly \enquote{translated} into its framework.

Furthermore, ParaDime uses neural networks to compute embeddings.
As a result, well-established loss functions from other tasks, such as classification and reconstruction, can be readily included in ParaDime DR routines.
This relates ParaDime to other techniques that add constraints to dimensionality reduction~\cite{vu_integrating_2022}.
In summary, ParaDime combines ideas from unifying nonlinear DR~\cite{bohm_attraction_2022,agrawal_minimum-distortion_2021} with parametric DR~\cite{van_der_maaten_learning_2009,sainburg_parametric_2021}, and provides flexibility to include alternative learning paradigms.

\section{The ParaDime Grammar of Parametric DR}
\label{sec:grammar}

The similarities between the various neighbor- and distance-based DR techniques outlined above inspired us to develop a unifying interface for specifying parametric dimensionality reduction \emph{routines}.
In ParaDime, routines are complete data processing pipelines that include all the specifications necessary to generate a trained parametric DR model from a given dataset.
In this section, we describe how routines can be specified with the ParaDime grammar of parametric DR.
This approach follows the tradition of grammars and grammar-like structures in the visualization community, such as Vega~\cite{satyanarayan_reactive_2016}, Vega-Lite~\cite{satyanarayan_vega-lite:_2017} and Encodable~\cite{wongsuphasawat_encodable_2020} for general visualizations, Atom~\cite{park_atom:_2018} for unit visualizations, Gosling~\cite{lyi_gosling_2022} for genome visualizations, and Neo~\cite{gortler_neo_2022} for confusion matrices.

\subsection{Overview}

ParaDime generalizes parametric DR by breaking it down into several steps, as outlined in the data-flow graph in Figure~\ref{fig:teaser}.
First, relations between all items in a given dataset are computed.
Then, a batch of data is sampled in a training loop.
The data batch is processed with a machine learning model, and new relations between all items in the processed batch are computed.
The batch-wise relations are compared with an appropriate subset of the overall relations to compute an embedding loss.
Additional losses may be added to the embedding loss.
Finally, the losses are used to optimize the machine learning model.

The ParaDime grammar defines how the building blocks for each of these steps are specified.
We use YAML for these specifications due to its focus on readability~\cite{dot_net_yaml_2021}.
A~ParaDime specification requires the three base-level entries  \gkey{relations}, \gkey{losses}, and \gkey{training phases}.
Additionally, the \gkey{derived data} field may be used to specify how extra data should be computed from the dataset or the relations. 
In the following subsections, we explain each of these fields in detail.
The model and dataset are not part of the specifications.
They are provided separately by the user, as explained in Section~\ref{sec:impl}.

\subsection{Relations}

The \gkey{relations} entry of a ParaDime specification lists \enquote{recipes} for computing mutual relations between data items.
Each relation recipe is specified either globally or at the batch level.
ParaDime computes \gval{global} relations between all items in the dataset before any training begins; these are typically relations between the original, high-dimensional data points.
In contrast, the computation of \gval{batch}-wise relations is deferred to the training-loop stage of the routine.
The \gval{batch}-wise relations are computed between items in a batch of data that has been processed by the model (i.e., between the low-dimensional data points).

A relation's \gkey{type} specifies how relations are computed; supported types are, for instance, exact pairwise distances (\gval{pdist}) and approximate neighbor-based distances (\gval{neighbor}).
A relation's \gkey{data} field specifies which part of the dataset to use to compute relations.
ParaDime assumes that individual parts of a dataset can be accessed via keys, which are used as values for the \gkey{data} field.
For example, a dataset might have its main data tensor and associated class labels stored under two different keys.
Relations typically accept a set of \gkey{options}.
For instance, distance-based relations allow users to specify the exact distance function to be used (e.g., \gkey{metric}: \gval{euclidean}).
Other relations allow algorithm-specific settings, such as the number of nearest neighbors for \gval{neighbor}-based relations.

Finally, a list of \gkey{transforms} can be applied to the relations.
Transforms can be used, for instance, to convert pairwise distances into perplexity-based probabilities of neighborhood as in \textit{t}-SNE~\cite{van_der_maaten_visualizing_2008} (see Section~\ref{sec:ex-tsne}).
The complete \gkey{relations} specification has the following structure:

\begin{spec}
~relations~:
  - ~name~: <rel name>
    ~level~: global | batch
    ~type~: <rel type>
    ~data~: <data field to use>
    ~options~: {...}
    ~transforms~:
      - ~type~: <transform type>
        ~options~: {...}
      - ...
  - ...
\end{spec}

Note that a routine can have any number of global or batch-wise relations.
Each relation has a name so that it can be referenced by the \gkey{losses} or in \gkey{derived data}.

\subsection{Losses}

Once ParaDime knows how to compute relations between data items, these relations can be used within \gkey{losses} to construct objective functions that govern the training process.
A~ParaDime specification of a routine's \gkey{losses} has the following structure:

\begin{spec}
~losses~:
  - ~name~: <loss name>
    ~type~: <loss type>
    ~func~: <loss function>
    ~keys~:
      ~data~: [<data attr name>, ...]
      ~rels~: [<rel name>, ...]
      ~methods~: [<model method>, ...]
  - ...
\end{spec}  

Each loss has a \gkey{type}, which defines how it behaves during training.
Supported loss types are \gval{relation}, \gval{classification}, \gval{reconstruction}, and \gval{position}.
A~loss of type \gval{relation} compares a subset of precomputed global relations to relations computed from a processed batch of data.
A~\gval{classification} loss compares the model output for a data batch to labels within the dataset.
A~\gval{reconstruction} loss compares the original input batch to an encoded and decoded version of the batch.
Finally, a~\gval{position} loss compares the low-dimensional output to a given set of coordinates.
To retain flexibility, each loss includes a specification of the \gkey{keys} that should be used to access the relevant model methods, attributes of the data, and/or the relations.
Losses can be combined during training to form weighted compound losses, as explained in the following subsection.

\subsection{Training Phases}

In ParaDime, the training of a routine is organized into \gkey{training phases}.
Each training phase consists of \gkey{sampling} and \gkey{optimizer} specifications, a number of \gkey{epochs}, and a \gkey{loss} specification:

\begin{spec}
~training phases~:
  - ~epochs~: <number of epochs>
    ~sampling~:
      ~type~: item | edge
      ~options~: {...}
    ~loss~:
      ~components~: [<loss name>, ...]
      ~weights~: [<weight>, ...]
    ~optimizer~:
      ~type~: <optim type>
      ~options~: {...}
  - ...
\end{spec}

The \gkey{sampling} type can be either \gval{item} (simple sampling of batches of items) or \gval{edge} (sampling of items based on relations between them).
The \gval{edge}-based sampling option enables ParaDime specifications of techniques that are based on negative-edge sampling~\cite{mikolov_distributed_2013,tang_visualizing_2016,mcinnes_umap:_2018} or triplets~\cite{chechik_large_2010} (see example in Section~\ref{sec:idea-supervised}).
As already mentioned above, the \gkey{loss} in each training phase is a weighted compound loss, whose \gkey{components} are specified with the names of the \gkey{losses} defined earlier.
Finally, the \gkey{optimizer} entry specifies which optimization technique to use (e.g., \gval{sgd}~\cite{lechevallier_large-scale_2010} and \gval{adam}~\cite{kingma_adam_2017}), along with options such as the learning rate or the momentum~\cite{sutskever_importance_2013}.

A~ParaDime routine can have any number of training phases.
Organizing the training into phases enables the pre-training of models, which can replace the initialization of low-dimensional positions used in non-parametric embeddings.
It also allows multi-stage optimization schemes such as the early exaggeration often used in \textit{t}-SNE~\cite{van_der_maaten_visualizing_2008}.

\subsection{Derived Data}

As mentioned earlier, an optional \gkey{derived data} field in a ParaDime specification allows new dataset attributes that are populated right before training to be defined based on other data attributes or on global relations.
They are specified as follows:

\begin{spec}
~derived data~:
  - ~name~: <attr name>
    ~data func~: <data function>
    ~keys~: [[data | rels, <key>], ...]
  - ...
\end{spec}

Here, the \gkey{keys} field allows users to specify which parts of the data or the relations are passed as arguments to the \gkey{data func} that computes the derived data.
A~simple use case for the \gkey{derived data} field would be the calculation of PCA for initialization purposes (see, e.g., the \textit{t}-SNE example in Section~\ref{sec:ex-tsne}).
Our rephrasing of parametric UMAP in terms of ParaDime in Section~\ref{sec:ex-umap} shows how derived entries can be used to set up initialization schemes based on transformed global relations.

\subsection{Using the Grammar}
\label{sec:impl}

ParaDime gives users two options for creating parametric DR routines.
The first option is to parse YAML specifications as described above.
In this case, users instantiate a ParaDime routine by loading a specification file and additionally passing a PyTorch~\cite{paszke_pytorch_2019} module as the model (i.e., neural network).
ParaDime then parses the specification and sets up Python objects corresponding to the components specified.
For each key in a specification, ParaDime allows only specific values that correspond to implemented classes or functions.
If users want to parse specifications with custom values, these values and the corresponding implementations need to be registered beforehand (using ParaDime's registration methods).
The second option is to set up the objects manually, using the ParaDime API rather than specification files.
In this case, custom objects and functions can be used directly.
Once users have instantiated a ParaDime routine, they can call its training method, passing the training data as an argument.
Since PyTorch modules are typically initialized randomly, most ParaDime routines constitute random embeddings until the training method is called.

We provide a detailed documentation with examples and a less technical introduction of the building blocks of ParaDime routines online~\cite{paradime_docs}.
Paradime is pip-installable, and the code is available on GitHub~\cite{paradime_github}.

\section{Framing Existing Techniques in Terms of ParaDime}
\label{sec:existing-techniques}

In this section, we show how (parametric extensions of) existing techniques can be specified in terms of the ParaDime grammar.
Note that we omit the \gkey{weights} list in all cases, as all examples use only a single loss component per training phase.

\subsection{Metric MDS}
\label{sec:ex-mds}

Metric multidimensional scaling aims to find a configuration of points in low-dimensional space such that the pairwise distances match those of the high-dimensional data~\cite{cunningham_linear_2015}.
This can be specified with ParaDime through Euclidean pairwise distance relations and a mean square error loss between the two relations:

\begin{spectc}
~relations~:
  - ~name~: dists hd
    ~level~: global
    ~type~: pairwise
    ~options~:
      ~metric~: euclidean
  - ~name~: dists ld
    ~level~: batch
    ~type~: pairwise
    ~options~:
      ~metric~: euclidean
~losses~:
  - ~name~: mds
    ~type~: relation
    ~func~: mse
    ~keys~:
      ~rels~:
        - dists hd
        - dists ld
~training phases~:
  - ~loss~:
      ~components~: mds
\end{spectc}

\begin{figure}
    \centering
    \includegraphics[width=\columnwidth]{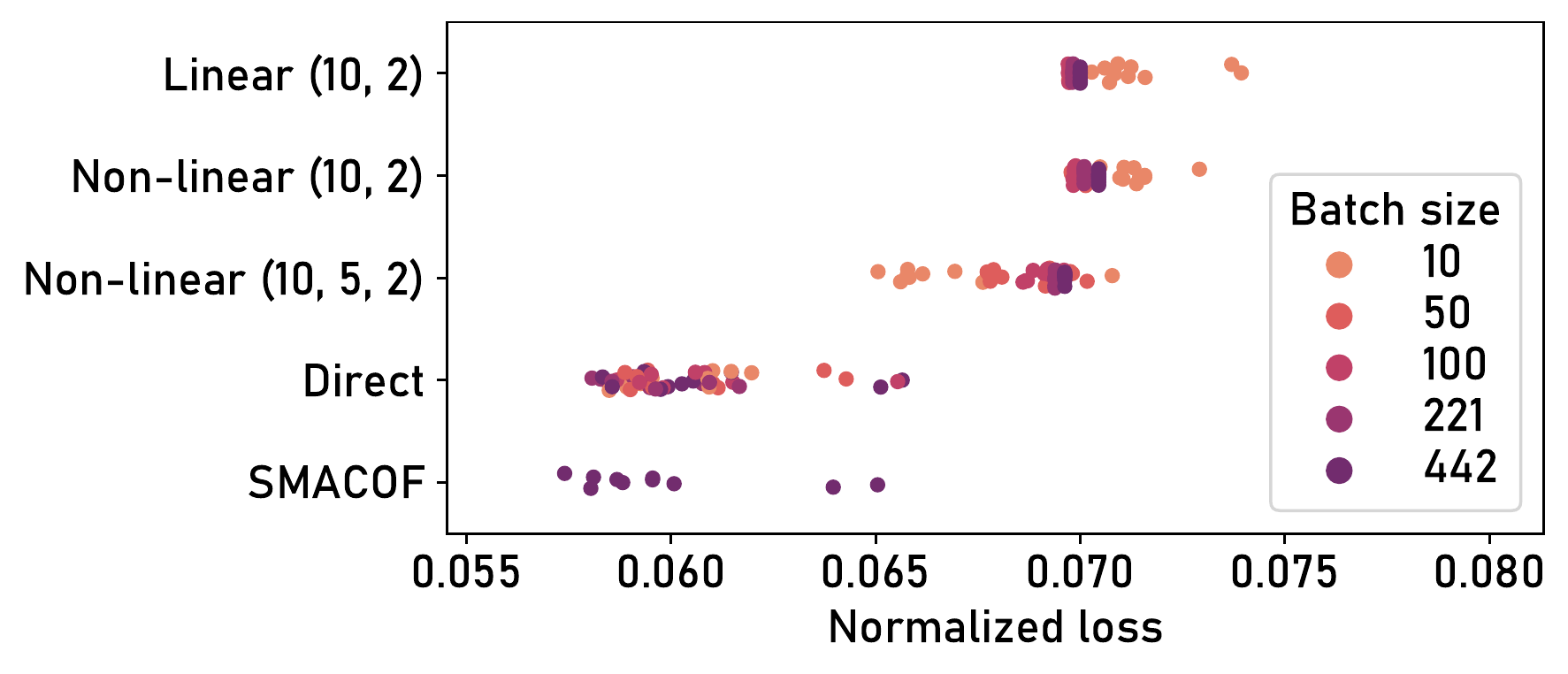}
    \caption{Normalized stress~\cite{espadoto_towards_2019} for parametric versions of metric MDS compared with the non-parametric SMACOF implementation of scikit-learn~\cite{pedregosa_scikit-learn:_2011}. The non-linear models were fully connected neural networks with hidden layer dimensions as indicated. The routine labeled \enquote{Direct} is a non-parametric routine using a batch-wise optimization which mimics that of the parametric ones. All models were trained on a 10-dimensional diabetes dataset with 442~items~\cite{efron_least_2004}.}
    \label{fig:mds}
\end{figure}

Figure~\ref{fig:mds} shows the normalized stresses~\cite{espadoto_towards_2019} for several ParaDime routines with different models and the specification above trained on a 10-dimensional diabetes dataset~\cite{efron_least_2004}.
The linear model was a simple matrix multiplication to map the 10-dimensional vectors to a 2-dimensional embeddings space.
The nonlinear models were fully connected neural networks with hidden layer dimensions as indicated, an additional bias, and softplus as activation function.
The routine labeled \emph{Direct} was a non-parametric routine implemented with ParaDime by replacing the model function with with a matrix that directly holds the embedding coordinates.
All ParaDime routines used the same optimizer (Adam~\cite{kingma_adam_2017}), learning rate (0.01) and number of epochs~(500).
The losses of the ParaDime routines are compared with that of the non-parametric scikit-learn implementation using the SMACOF algorithm~\cite{kruskal_multidimensional_1964}.
Note how the routines with linear and nonlinear models of size \(10\times2\) performed almost identically.
Adding another hidden layer of dimension five reduced the loss substantially, especially for smaller batch sizes.
The average loss for a batch size of ten was less than \SI{12}{\percent} greater than the average of the SMACOF baseline, despite the simplicity of the model and the absence of hyperparameter tuning.
Interestingly, for the two models of size \(10\times2\), smaller batch sizes led to higher losses.
The non-parametric implementation had losses similar to the SMACOF baseline.
These results reveal the importance of model and hyperparameter selection, which we discuss in Section~\ref{sec:discussion-limitations}.

\subsection{\textit{t}-SNE}
\label{sec:ex-tsne}

The \textit{t}-SNE algorithm begins with calculating pairwise distances that are transformed into normalized and symmetrized probabilities of high-dimensional neighborhood based on a perplexity hyperparameter~\cite{van_der_maaten_visualizing_2008}.
In low-dimensional space, probabilities of neighborhood are calculated by transforming Euclidean distances with a Student's t-distribution~\cite{van_der_maaten_visualizing_2008}.
Defining these two relations in ParaDime using \gkey{transforms} is straightforward.
Note that the global relation specification contains \gval{neighbor} rather than \gval{pdist} as \gkey{type}, which tells ParaDime to use approximate nearest-neighbor-based distances.
This is an optimization that is used in modern \textit{t}-SNE implementations~\cite{policar_opentsne:_2019}.
The two probability matrices are compared using the KL divergence.

Before this step, most \textit{t}-SNE implementations perform an initialization of the embedding with PCA coordinates.
The embedding coordinates, however, cannot be initialized directly in a parametric DR routine, because the coordinates are outputs of a neural network.
Instead, the model weights have to be set in such a way that the model mimics a PCA transformation.
In ParaDime, this is achieved by pre-training the model in a separate training phase.
The \gkey{derived data} specification makes the required PCA coordinates available during training.
This results in the following ParaDime specification for parametric \textit{t}-SNE:

\begin{spectc}
~derived data~:
  - ~name~: pca
    ~data func~: pca
    ~keys~: [[data, main]]
~relations~:
  - ~name~: p
    ~level~: global
    ~type~: neighbor
    ~data~: main
    ~options~:
      ~metric~: euclidean
    ~transforms~:
      - ~type~: perplexity
        ~options~:
          ~perplexity~: <p>
      - ~type~: symmetrize
      - ~type~: normalize
  - ~name~: q
    ~level~: batch
    ~type~: pairwise
    ~data~: main
    ~options~:
      ~metric~: euclidean
    ~transforms~:
      - ~type~: t-dist
        ~option~s:
          ~alpha~: 1.
      - ~type~: normalize
~losses~:
  - ~name~: init
    ~type~: position
    ~func~: mse
    ~keys~:
      ~data~: [main, pca]
  - ~name~: emb
    ~type~: relation
    ~func~: kl div
    ~keys~:
      ~rels~: [p, q]
~training phases~:
  # pca initialization
  - ~loss~:
      ~components~: [init]
      ~sampling~:
        ~type~: item
  # main embedding
  - ~loss~:
      ~components~: [emb]
      ~sampling~:
        ~type~: item
\end{spectc}

Parametric \textit{t}-SNE as specified above does not feature early exaggeration~\cite{van_der_maaten_visualizing_2008}.
However, this can easily be implemented by adding a training phase between the pre-training and embedding phases, making use of a simple multiplicative transform.
In contrast to the parametric version of \textit{t}-SNE recently introduced by Lai et al.~\cite{lai_parametric_2022}, ParaDime currently does not use gradient clipping.
In future version, gradient clipping could be included as an option in the loss specification.

An example of a parametric \textit{t}-SNE routine implemented with ParaDime is shown in the right part of Figure~\ref{fig:teaser}.
It was trained on a subset of 5000 images of the MNIST dataset of handwritten digits~\cite{lecun_mnist_2005} with a perplexity of 100 and a learning rate of 0.001.
The model had hidden layer dimensions of 1024, 512, 256, 128 and used softplus for all activation functions.
This model architecture is the same as the one used by Lai et al.~\cite{lai_parametric_2022}, but our experiments suggest that models with far fewer parameters (e.g., hidden layer dimensions of 100 and 50) work reasonably well in many cases.
Figure~\ref{fig:teaser} also shows the result of applying the trained model to 15,000 unseen data instances.

\subsection{UMAP}
\label{sec:ex-umap}

As discussed in Section~\ref{sec:related-work}, UMAP has several conceptual similarities to \textit{t}-SNE.
Its ParaDime specification therefore reads relatively similar to that of \textit{t}-SNE.
In the following, we omitted fields that are the same as in the specification for parametric \textit{t}-SNE.

\begin{spectc}
~derived data~:
  - ...
  - ~name~: spectral
    ~data func~: spectral
    ~keys~: [[rels, p]]
~relations~:
  - ...
    ~transforms~:
      - ~type~: connect
        ~options~:
          ~neighbors~: <n>
      - ~type~: symmetrize
        ~options~:
          ~sub prod~: true
      - ~type~: normalize
  - ...
    ~transforms~:
      - ~type~: cauchy
        ~options~:
          ~spread~: <s>
          ~min dist~: <md>`\pagebreak`
~losses~:
  - ~name~: init
    ~type~: position
    ~func~: mse
    ~keys~:
      ~data~:
        - main
        - spectral
  - ~name~: emb
    ~type~: relation
    ~func~: cross entropy
    ~keys~:
      ~rels~: [p, q]
~training phases~:
  # spectral init
  - ...
  # main embedding
  - ~loss~:
      ~components~: [emb]
      ~sampling~:
        ~type~: edge
\end{spectc}

Sainburg et al.~\cite{sainburg_parametric_2021} outlined the main differences of UMAP from \textit{t}-SNE. In contrast to \textit{t}-SNE, UMAP:
\begin{itemize}
    \item initializes coordinates with a \gval{spectral} embedding based on global relations, instead of applying PCA;
    \item transforms distances to probabilities with kernels whose widths depend on \gval{connectivity} instead of perplexity;
    \item transforms batch-wise relations with a modified \gval{cauchy} distribution instead of a Student's t-distribution;
    \item uses \gval{cross entropy} as loss instead of KL divergence; and
    \item uses negative-\gval{edge} sampling instead of item-based sampling.
\end{itemize}
ParaDime uses an implementation of negative-edge sampling which does not ensure that each item is sampled at least once.
This may lead to slightly smaller repulsive forces in ParaDime embeddings compared to an existing parametric UMAP version~\cite{sainburg_parametric_2021}.

The bottom four scatterplots in Figure~\ref{fig:idea-hybrid} give an indication of how parametric UMAP embeddings look for the MNIST dataset~\cite{lecun_mnist_2005}.
Note, however, that these embeddings come from routines with an additional loss term, as explained in Section~\ref{sec:idea-hybrid}.

\subsection{Additional Neighbor-based Techniques}
\label{sec:neighbor-tech}

ParaDime includes implementations of all \gkey{relations}, \gkey{transforms}, and \gkey{data func} methods specified in the examples above.
With these methods, it is also possible to specify LargeVis~\cite{tang_visualizing_2016}, which basically combines \textit{t}-SNE's high-dimensional relations with negative-edge sampling.
LargeVis is not restricted to a specific transform for the low-dimensional (i.e., \gval{batch}-wise) relations; the authors state that \enquote{many probabilistic functions can be used} instead~\cite{tang_visualizing_2016}.
This aligns well with ParaDime's flexible concept of transforms.

Isomap is another neighbor-based technique, but it uses geodesic distances instead of probabilities of neighborhood~\cite{tenenbaum_global_2000}.
Specifying Isomap with ParaDime merely requires implementing either a new \gval{relations} type or a \gval{transform} that converts Euclidean distances to geodesic distances.

\subsection{Classifiers \& Autoencoders}
\label{sec:ex-classifier}

In addition to the \gval{relation}-type loss used in all DR techniques discussed so far, ParaDime also provides losses for typical machine-learning tasks that are not limited to DR.
In particular, the \gval{classification} loss makes it straightforward to implement classification models.
The following specification assumes that the main data is accessible as \gval{main}, and ground truth labels as \gval{labels}.

\begin{spec}
~losses~:
  - ~type~: classification
    ~func~: cross entropy
    ~keys~:
      ~data~:
        - main
        - labels
\end{spec}

Similarly, autoencoders can be concisely specified using the predefined \gval{reconstruction} loss.
Graving and Couzin~\cite{graving_vae-sne_2020}, and Sainburg et al.~\cite{sainburg_parametric_2021} have previously discussed the potential of combining the reconstruction ability of autoencoders with relation-based embedding losses.

\section{Experimenting with Combined Techniques}
\label{sec:new-ideas}

In this section, we present several application ideas for ParaDime.
These examples show the versatility of the ParaDime specifications, and encourage experimentation with new ideas that emerge from combining different losses.

\begin{figure*}[t]
    \includegraphics[width=\textwidth]{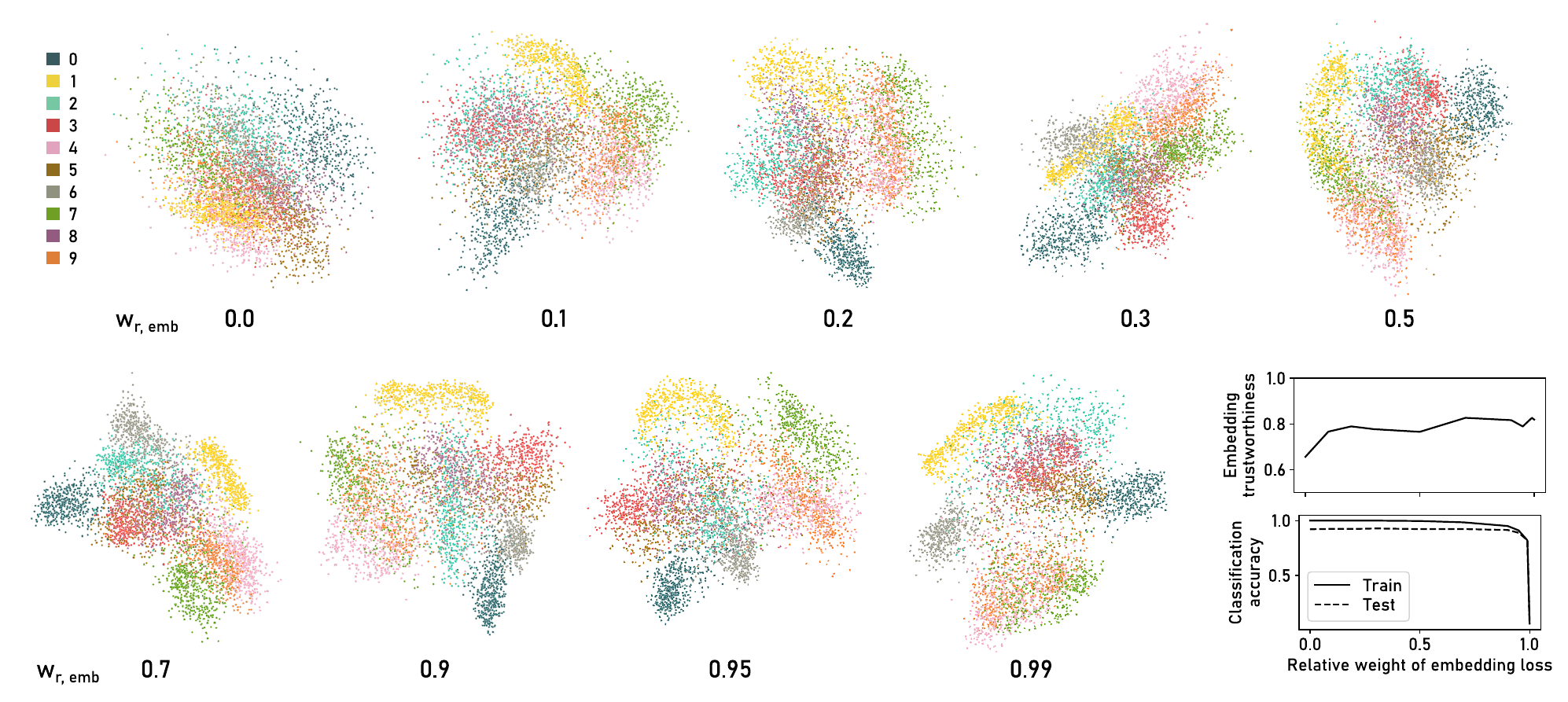}
    \caption{Embeddings of hybrid embedding\slash classification routines for the MNIST dataset~\cite{lecun_mnist_2005} created with ParaDime. The relative weight of the embedding loss component is indicated by \(w_{r, \text{emb}}\), and the weight of the classification component was \(1 - w_{r, \text{emb}}\). All embedding-related specifications were the same as those of the ParaDime parametric UMAP routine. The routines were trained on a subset of 5000 randomly sampled MNIST images. Test accuracy was calculated on a different subset of 5000 images. Trustworthiness~\cite{goos_neighborhood_2001,espadoto_towards_2019} was calculated based on ten nearest neighbors.}
    \label{fig:idea-hybrid}
\end{figure*}

\subsection{Hybrid UMAP for Embedding and Classification}
\label{sec:idea-hybrid}

In Sections~\ref{sec:ex-umap} and \ref{sec:ex-classifier} we showed how to use ParaDime to specify a parametric version of UMAP and a simple classification model, respectively.
In this section, we combine the two to create a hybrid embedding and classification routine which uses a shared latent space for both tasks.
We applied our multitask routine to the MNIST dataset of handwritten digits~\cite{lecun_mnist_2005}.

As a model, we used a fully connected network with hidden-layer dimensions 100 and~50.
The model has two output layers: one of dimension ten that yields the logits used for classification, and one of dimension two for the embedding.
Both these output layers are connected to the second hidden layer.

As explained above, UMAP uses edge-based sampling.
When edge-based sampling is specified in ParaDime, each batch contains not only the pairs of vertices between the sampled edges, but also a list of unique data items suitable for other tasks, such as classification.
Therefore, losses that require item-based sampling can readily be added to routines that use negative-edge sampling.
The specification below  creates our hybrid classification and embedding model, with previously defined losses and relations omitted.

\begin{spec}
~relations~: <UMAP relation specs>
~losses~: [<UMAP loss>, <classification loss>]
~training phases~:
  - ~loss~:
      ~components~: [umap, class]
      ~weights~: <w>
\end{spec}

Thanks to ParaDime's specification interface, the losses above can be simply reused as components in a compound loss.
Figure~\ref{fig:idea-hybrid} shows nine embeddings created with different weights for the loss components.
All routines were trained on the same subset of 5000 images from MNIST for 100 epochs and without any pre-training.
Figure~\ref{fig:idea-hybrid} also includes plots of the classification accuracy and the embedding trustworthiness (as defined by Venna and Kaski~\cite{goos_neighborhood_2001,espadoto_towards_2019}) as functions of the weight.
The accuracy was calculated using a non-overlapping test subset of 5000 random images.
Note that even a small weight on the embedding loss leads to a substantial class separation in the scatterplots.
At the same time, classification accuracy is not affected by the additional embedding task.
The accuracy suffers only when the weight on the classification approaches zero.
Weighting the embedding with values in the wide range of 0.5~to~0.95 produces visually \enquote{sensible} embeddings with relatively high trustworthiness and practically the same classification accuracy as the pure classifier.
In fact, some of our experiments showed that the additional embedding loss can slightly improve generalization of the classifier.
This observation is in line with the original motivation for multitask learning~\cite{caruana_multitask_1997}.

Such a hybrid embedding and classification model could form the basis for a visualization tool in which users can add new points to existing embeddings.
The predicted class labels could be used to visually encode the new data points and/or to inform users whether a new point lies within a region of the embedding where other points of the same class are located.

\subsection{Supervised \textit{t}-SNE with Triplet Loss}
\label{sec:idea-supervised}

\begin{figure*}[t]
    \includegraphics[width=\textwidth]{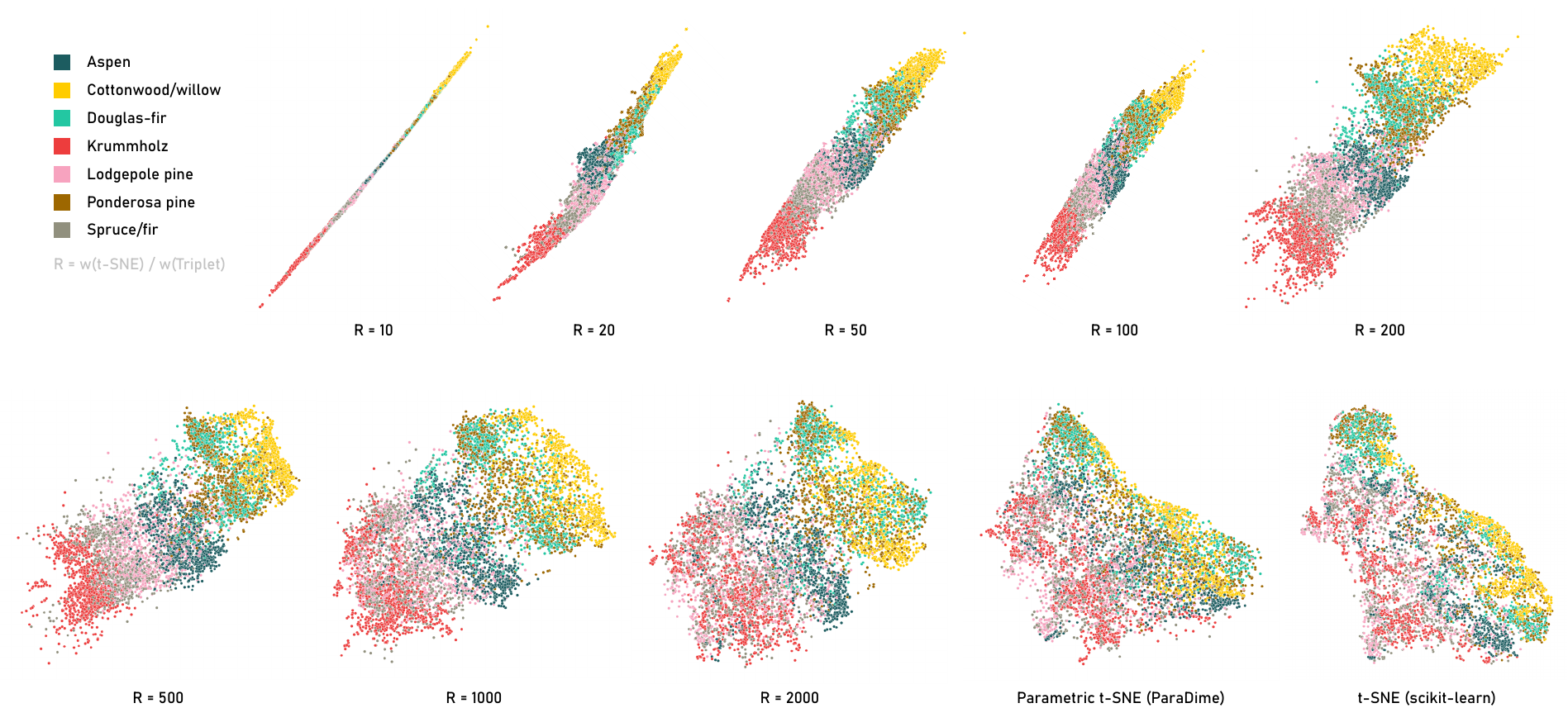}
    \caption{Supervised embeddings of a subset of the forest covertype dataset~\cite{chechik_large_2010}. All embeddings labeled with \(R\) are supervised versions of parametric \textit{t}-SNE, where supervision was included by means of a triplet loss based on the ground truth labels. \(R\) is the ratio of the weights of the \textit{t}-SNE loss and the triplet loss. For comparison, embeddings created with scikit-learn's non-parametric \textit{t}-SNE implementation and with a plain ParaDime \textit{t}-SNE version (using item-based sampling and no triplet loss) are shown. The perplexity was 200 in all cases, and a class-balanced subset of 7000 items was used.}
    \label{fig:idea-supervised-triplet}
\end{figure*}

In this example, we combined a parametric version of \textit{t}-SNE (see Section~\ref{sec:ex-tsne}) with a triplet loss~\cite{chechik_large_2010} to learn several supervised embeddings for the forest covertype dataset~\cite{blackard_comparative_1999}.
This is an example of an instance-level constraint as categorized by Vu et al.~\cite{vu_integrating_2022}.

The forest covertype dataset consists of \num{581012} records with 54 attributes each.
Each item corresponds to a \(\SI{30}{\meter}\times\SI{30}{\meter}\) cell of a US region, and the attributes describe cartographic variables, such as elevation, slope, and distance from the nearest roadway.
Each item is labeled with the ground truth value for the type of trees covering the cell (e.g., aspen, krummholz, and spruce/fir).
The dataset is strongly imbalanced, with the most prevalent class being more than 100 times more frequent than the least.
In this example, we used the first ten numerical attributes and sampled an almost balanced subset of 7000~items.

Supervising \textit{t}-SNE with an additional term based on triplets can be achieved easily thanks to the ParaDime interface:
First, we use negative-edge sampling to construct triplets. 
In negative-edge sampling,  rather than batches of individual items, batches of edges between items are sampled during training.
In other applications of this sampling strategy (e.g., UMAP~\cite{mcinnes_umap:_2018}), a positive edge is sampled according to the probabilities of neighborhood of the two points (i.e., vertices).
A~specified number of random negative edges for one of the two vertices is then added.
Negative edges are edges between two vertices for which the probability of neighborhood is zero.
In this example, we instead created a probability matrix \(r\) with \(r_{ij}=1\) if \(g_i = g_j\) and 0 else, where \(g_i\) are the ground truth labels of the data.
If we use this probability matrix for negative-edge sampling with a negative sampling rate of one, we essentially sample one pair of vertices \((a, b)\) with equal labels and another pair \((a,c)\) with different labels.
The set of vertices \(a,b,c\) constitutes a triplet~\cite{chechik_large_2010, balntas_learning_2016}.
We can then simply add an additional triplet loss.
This results in the following ParaDime specification:

\begin{spectc}
~derived data~: [<PCA>]
~relations~:
  - <global t-SNE rel>
  - <batch t-SNE rel>
  - ~name~: r
    ~level~: global
    ~type~: pairwise eq
~losses~:
  - <PCA init loss>
  - <t-SNE loss>
  - ~name~: triplet
    ~type~: triplet
    ~func~: margin
    ~keys~:
      ~data~: [main, data]
~training phases~:
  - <PCA init>
  - ~loss~:
      ~components~:
        - tsne
        - triplet
      ~weights~: <w>
      ~sampling~:
        ~type~: edge
        ~options~:
          ~rels~: r
          ~rate~: 1
\end{spectc}

Here, \gval{pairwise eq} stands for the global relation as defined by \(r_{ij}\), and \gval{margin} is the name of the following loss function that is applied to the triplets~\cite{wang_learning_2014,balntas_learning_2016}:
\begin{equation}
    L_\text{triplet}(a,b,c) = \max(d(a,b)-d(a,c)+m, 0),
\end{equation}
where \(m\) is the margin hyperparameter.
We abridged the parts of the specification that match that of \textit{t}-SNE from Section~\ref{sec:ex-tsne}.

Figure~\ref{fig:idea-supervised-triplet} shows eight versions of embeddings specified this way, with different values for the loss weights.
In all cases, the model was a fully connected neural network with hidden layer dimensions 100 and 50.
Each embedding was initialized with a PCA-based pre-training for ten epochs with item sampling and a batch size of 500.
As explained above, the main embedding phases used negative-edge sampling, with 300 triplets being sampled in each batch.
For comparison, Figure~\ref{fig:idea-supervised-triplet} includes a parametric \textit{t}-SNE without the extra triplet loss and with regular item sampling.
We also show the result of scikit-learn's non-parametric \textit{t}-SNE.
For all embeddings the perplexity value was set to 200.

For the triplet loss as defined above to be minimal, the distance along negative edges (i.e., between a pair of items with different labels) must be substantially larger than the distances along a positive edge.
This pulls together items from the same class.
Putting too much weight onto the triplet loss causes all items to condense along a single line, approximately sorted by their class labels.
As the weight of the triplet loss is reduced, the structure of the \enquote{pure} \textit{t}-SNE is increasingly preserved, while classes are well separated (see, e.g., the embeddings for \textit{t}-SNE\slash triplet loss weight ratios of 1000 in Figure~\ref{fig:idea-supervised-triplet}).
With vanishing weight on the triplet loss, the embedding still differs noticeably from that which used item-based sampling; here, the triplet sampling strategy might be disadvantageous, as it favors certain batch configurations over others.

One potential application idea for such supervised embeddings is an interactive visual interface for dataset exploration, that allows users to switch between a purely attribute-driven visualization (e.g., pure \textit{t}-SNE) and a supervised one with more pronounced class separation.
In the former, users could explore similarities and differences between all data points as usual, while the latter would enable class-specific exploration without losing track of the overall structure.

\subsection{Attribute-guided Embeddings}

\begin{figure*}[t]
    \includegraphics[width=\textwidth]{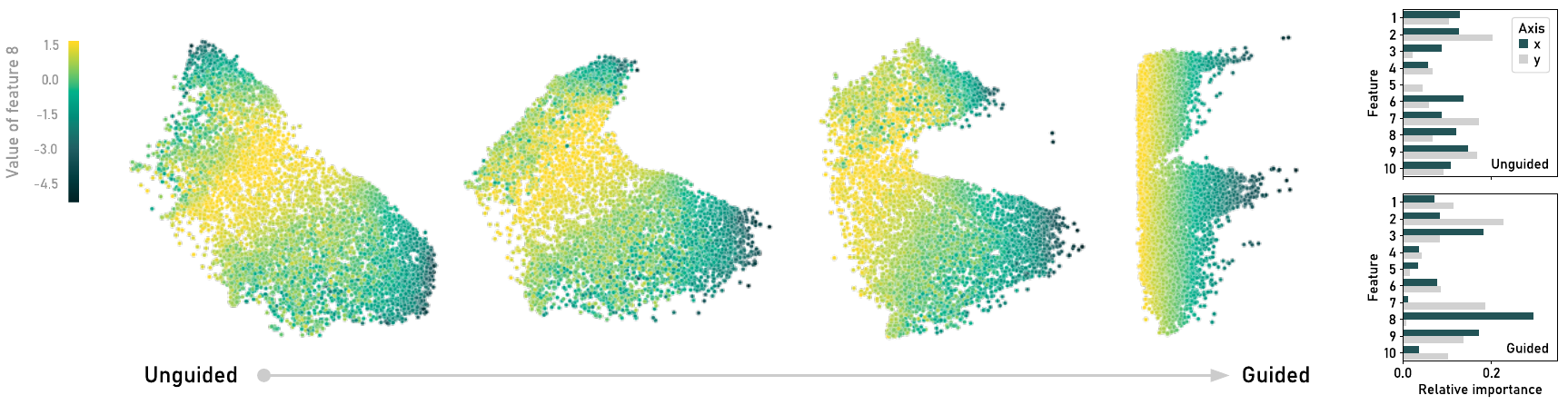}
    \caption{Attribute-guided embeddings of a subset of the forest covertype dataset~\cite{chechik_large_2010}. Attribute guiding was implemented by combining \textit{t}-SNE with a correlation loss which orders the data points along the \(x\)-axis by the value of the eighth feature (hillshade at noon).
    The weights for the embeddings shown are \((w_\text{\textit{t}-SNE}, w_\text{corr}) = (1,0)\), \((5000, 1)\), \((1000, 1)\), and \((100, 1)\), respectively. The bar chart on the right shows, based on integrated gradients, the feature importance scores for the learned embeddings.}
    \label{fig:idea-attribute-guided}
\end{figure*}

In this example, we again look at embeddings of the covertype dataset discussed in the previous section.
This time, however, our primary interest is not in the class distribution, but in using specific attributes to guide the embeddings.
In particular, we used ParaDime to construct an embedding in which a specified direction correlates with one of the high-dimensional attributes.
To this end, we defined a new type of loss:
\begin{equation}
    L_\text{corr}(a, b; i, j) = 1 - \biggl( \frac{\cov(a_i, b_j)}{\sigma_{a_i}\sigma_{b_j}} \biggr)^2.
    \label{eq:corr-loss}
\end{equation}
Here, \(a\) and \(b\) are two data matrices with the same number of rows, and \(a_i\) and \(b_j\) refer to columns \(i\) and \(j\), respectively; \(\cov\) is the covariance, and \(\sigma\) is the standard deviation.
This loss is equivalent to one minus the squared Pearson's correlation coefficient for the \(i\)th column of \(a\) and the \(j\)th column of \(b\).
During the training of our routine, \(a\) will be a batch of high-dimensional data and \(b\) the processed (i.e., embedded) 2-dimensional batch.

Having defined a loss \gval{corr} that uses the function \(L_\text{corr}\) (Eq.~\ref{eq:corr-loss}) and applies it to the unprocessed and embedded versions of the input batch, we can simply construct a compound loss analogously to the other examples in this section.
The loss components (\textit{t}-SNE loss and correlation loss) can be weighted, and the dimensions that should correlate can be specified as options to the loss.

Figure~\ref{fig:idea-attribute-guided} shows four examples of such attribute-guided embeddings with different weights.
In all examples, \(i\) was set to eight and \(j\) to one, which means that the \emph{Hillshade (noon)} attribute of the covertype dataset was constrained to correlate with the \(x\)-direction of the embedding.
In the embeddings in Figure~\ref{fig:idea-attribute-guided}, the points are colored by the high-dimensional attribute value specified.
With increasing weight on the correlation loss, the embedding is distorted such that the values decrease from left to right, while the remaining structure is preserved to some extent.
Within a certain range of weights, the transition from unguided to strongly guided embeddings appears to be smooth, with the points \enquote{folding over} continuously to satisfy the constraints.

Because ParaDime models are neural networks, we can apply to them any existing explanation technique developed for neural networks.
In this example, we sought to verify that the attribute we specified (feature eight, \emph{Hillshade (noon)}) was actually of high importance for the resulting \(x\) value.
To this end, we applied a \enquote{vanilla} version of integrated gradients~\cite{molnar_interpretable_2022} to our model.
The resulting feature importance scores are shown in the bar chart in Figure~\ref{fig:idea-attribute-guided}.
Note that for the strongly guided embedding, feature eight is indeed the most important for the \(x\) result by some margin, and it does not contribute to \(y\) at all.

Attribute-guided embeddings are not only a showcase for how easily new techniques can be constructed with ParaDime.
They might be useful in cases in which users want to transition from purely unsupervised embeddings to ones where a specified attribute is of particular interest to the analysis.

\section{Discussion}
\label{sec:discussion}

In this section, we discuss some of the design choices related to the structure of the ParaDime grammar and its implementation.
We also reflect on ParaDime's ease of use, its customizability, limitations, and future work.

\subsection{Structure of the Grammer}
\label{sec:discussion-grammar}

The structure of the ParaDime grammar cannot be uniquely derived from the necessary building blocks (dataset, relations, etc.), but depends on a number of choices.
For example, in an earlier version of the specifications, losses were defined entirely within the training phases, and their specification included a weight.
However, this strongly limited the reusabilty of losses across phases.
We thus opted for loss specifications at the base level, which required the introduction of the \gkey{components} and \gkey{weights} entries, and the use of loss names that could be referenced.

Furthermore, we initially planned different base-level entries for lists of global and batch-wise relations.
From a computational view, they are typically used at different times in the routines, and only the batch-wise relations must be differentiable.
Nevertheless, we ultimately chose a flat list of relations with individual \gkey{level} entries to highlight the conceptual similarities between them.

Initially, we had also planned to include a \gkey{model} entry in the ParaDime specifications.
Our first draft included a nested structure of (sub-)model specifications based closely on how PyTorch allows arbitrarily nested modules.
However, we soon realized that the creation of a general declarative grammar for neural networks went well beyond the scope of this work.
We thus decided to have users pass their PyTorch module to ParaDime alongside a DR specification.
ParaDime can also construct a default, fully connected model to help users to get started.

\subsection{Implementation Choices}
\label{sec:discussion-implementation}

We considered PyTorch~\cite{paszke_pytorch_2019}, TensorFlow~\cite{abadi_tensorflow:_2016}, and JAX~\cite{frostig_compiling_2018} as machine learning frameworks for ParaDime.
Ultimately, we settled on PyTorch because it has become the most popular framework for research purposes~\cite{pytorch-vs-tensorflow}.

While in this work we used YAML~\cite{dot_net_yaml_2021} for the specifications in this paper due to its focus on readability, ParaDime is also capable of parsing JSON specifications with the same structures.
In addition to construction by specifications (which facilitate sharing and reproducibility) ParaDime allows an object-oriented construction of routines, as this is particularly suitable for adapting existing routines or dynamically changing properties of routines.

\subsection{Ease of Use and Customization}

As outlined in Section~\ref{sec:neighbor-tech}, ParaDime can be readily used for distance- and neighbor-based DR techniques.
We asked an AI Bachelor student with no prior deep learning experience to implement a parametric version of Isomap using ParaDime.
Without ParaDime, the student would have had to write a sampling routine that correctly incorporates pairwise relations, set up the PyTorch module, write and correctly apply the embedding loss, and set up the optimization loops.
With ParaDime, the student only had to wrap code for the geodesic distance computation (taken from scikit-learn) in a ParaDime relations class.
Because ParaDime already offers differentiable implementations of several batch-wise relations, the student did not have to learn PyTorch at all.

For more obscure DR techniques, users must program custom losses or batch-wise relations.
Ensuring that all relevant parts remain differentiable requires some understanding of PyTorch.
Currently, the sampling procedure is the most difficult part of the routines to customize, since it is not directly accessible through the ParaDime API.
However, we believe that the built-in item- and edge-based samplers should suffice for most cases.
Even in highly customized applications, ParaDime should reduce overhead because it takes care of most of the data handling, facilitates the combination of multiple losses, and/or sets up the training loops.

\subsection{Limitations \& Future Work}
\label{sec:discussion-limitations}

One major limitation when moving from traditional DR techniques to parametric embeddings is the increased number of hyperparameters.
Users must select a suitable model architecture and set batch sizes, optimizers and learning rates such that the loss is properly minimized.
For the predefined ParaDime routines, we provide defaults based on our own experiments.
With new routines, however, finding suitable choices for hyperparameters can be challenging.
The same is true for weights in compound losses.
Choosing suitable weights for the loss components is a long-standing problem in multitask learning~\cite{gong_comparison_2019}.
As a result, non-obvious weight ratios have to be tried out, as seen in some of the examples discussed in Section~\ref{sec:new-ideas}.
However, ParaDime's focus on reusability and ease of specification facilitates experiments with different weights.
ParaDime also features built-in plotting utilities, which allow users to rapidly check the embeddings visually.

Another limitation related to batch-wise training is that certain global constraints are difficult to implement.
For example, global density-based measures such as that used in densMAP~\cite{narayan_assessing_2021} are challenging to reproduce from small batches.
In principle, the batch size in ParaDime can be set to the number of items in the dataset to allow computation of global measures during training.
However, this might lead to problems with gradients for other losses.
We plan to experiment with such globally constrained techniques to provide better ways of incorporating them.

Finally, we plan to include export utilities for the trained models so that they can easily be used elsewhere.
It would be particularly desirable to export models in a format that could be used directly within a web-browser.
Visualizations implemented as web-apps could thus make use of pre-trained ParaDime routines without the need for a backend.

\section{Conclusion}
\label{sec:conclusion}

We have introduced ParaDime, a framework for parametric dimensionality reduction.
The ParaDime grammar allows users to specify DR routines in a declarative way.
We have shown how this approach enables parametric extension of existing techniques and illustrated how ParaDime facilitates experimentation with new ideas.
We hope that---due to our focus on flexibility and customization---ParaDime will inspire further research into the potential of parametric dimensionality reduction.

\section*{Acknowledgements}

This work was supported by the State of Upper Austria and the Austrian Federal Ministry of Education, Science and Research via the Linz Institute of Technology (LIT-2019-7-SEE-117), the State of Upper Austria (Human-Interpretable Machine Learning), the Austrian Science Fund (FWF DFH 23--N), and the Austrian Research Promotion Agency (FFG 881844). Pro\textsuperscript{2}Future is funded within the Austrian COMET Program under the auspices of the Austrian Federal Ministry for Climate Action, Environment, Energy, Mobility, Innovation and Technology, the Austrian Federal Ministry for Digital and Economic Affairs, and of the States of Upper Austria and Styria. COMET is managed by the Austrian Research Promotion Agency FFG.

\printbibliography

\end{document}